\documentclass{bmvc2k}

\usepackage{amsfonts}
\usepackage{dsfont}
\usepackage{booktabs}
\usepackage{caption}
\usepackage{indentfirst} 
\usepackage[capitalize]{cleveref}
\crefname{section}{Sec.}{Secs.}
\Crefname{section}{Section}{Sections}
\Crefname{table}{Table}{Tables}
\crefname{table}{Tab.}{Tabs.}

\usepackage{wrapfig}

\makeatletter
\DeclareRobustCommand\onedot{\futurelet\@let@token\@onedot}
\def\@onedot{\ifx\@let@token.\else.\null\fi\xspace}

\def\eg{\emph{e.g}\onedot} 
\def\ie{\emph{i.e}\onedot} 
 
 \def\vs{\emph{vs}\onedot}

\makeatother

\usepackage[ruled,vlined]{algorithm2e}

\definecolor{commentcolor}{RGB}{110,154,155}   %
\newcommand{\PyComment}[1]{\ttfamily\textcolor{commentcolor}{\# #1}}  %
\newcommand{\PyCode}[1]{\ttfamily\textcolor{black}{#1}} %

\newcommand{\modelname}{ITGC\@\xspace}

\usepackage{tcolorbox}
\definecolor{promptcolor}{RGB}{242, 242, 240}

\usepackage{listings}

\title{Interpretable Text-Guided Image Clustering via Iterative Search}

\addauthor{Bingchen Zhao}{zhaobc.gm@gmail.com}{1}
\addauthor{Oisin Mac Aodha}{oisin.macaodha@ed.ac.uk}{1}

\addinstitution{
 University of Edinburgh\\
 Edinburgh, UK
}

\runninghead{Zhao and Mac Aodha}{ITGC}

\def\eg{\emph{e.g}\bmvaOneDot}

\begin{document}

\maketitle
\begin{center}
\centering
\captionsetup{type=figure}
    \includegraphics[width=0.8\linewidth]{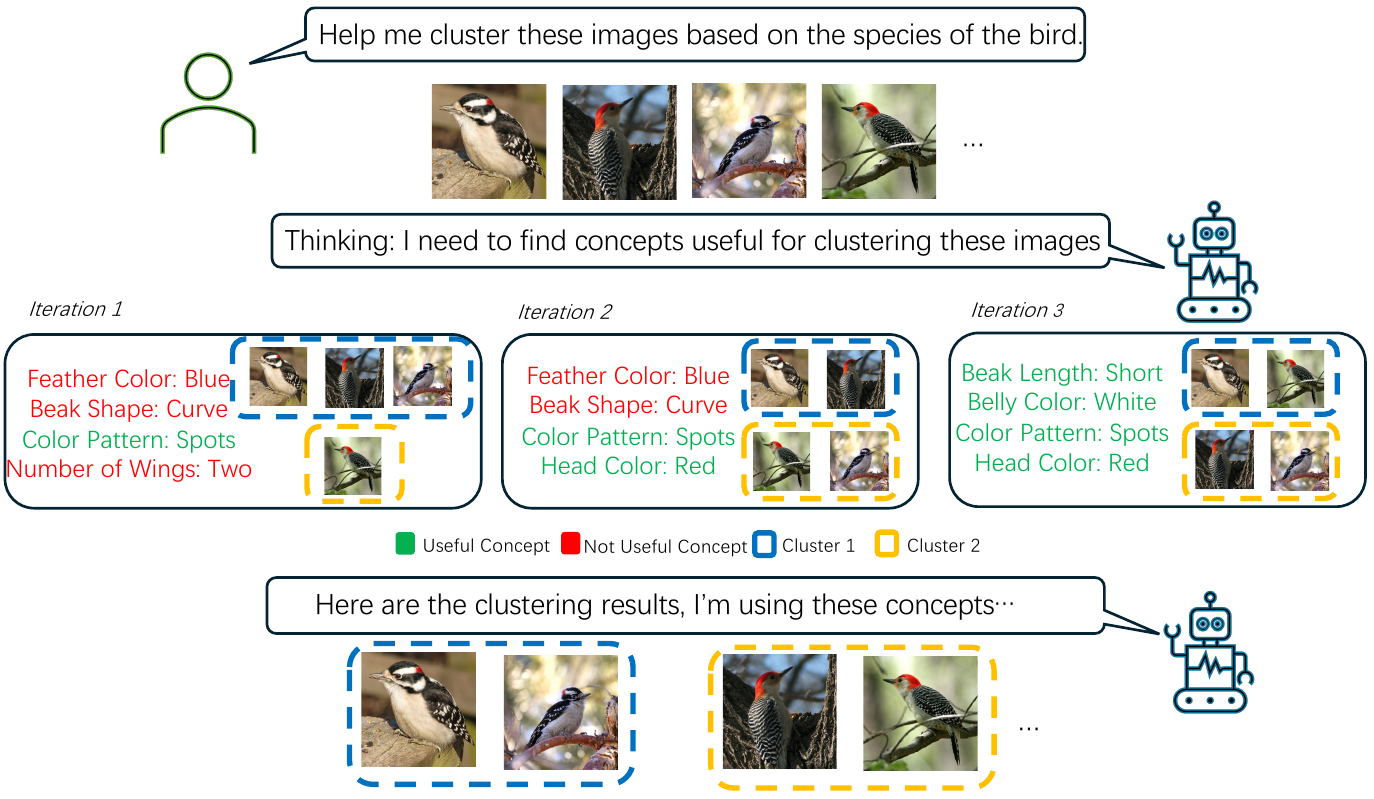}
\vspace{5pt}
\caption{ 
We propose \textbf{\modelname}, a new approach for text-guided image clustering that takes a set of unlabeled images, along with  a high-level user text query, as input and automatically returns a clustering of the images that is aligned to the user's intent. 
\modelname conducts an iterative search procedure, guided by an unsupervised  objective, that enables it to discover interpretable visual concepts to assist clustering. 
}
\vspace{-3pt}
\label{fig:teaser}
\end{center}

\begin{abstract}
Traditional clustering methods aim to group unlabeled data points based on their similarity to each other. 
However, clustering, in the absence of additional information, is an ill-posed problem as there can be many different, yet equally valid, ways to partition a dataset. 
Distinct users may want to use different criteria  to form clusters in the same data, \eg shape \vs color. 
Recently introduced text-guided image clustering methods aim to address this ambiguity by allowing users to specify the clustering criteria of interest using natural language instructions. 
This instruction provides the necessary context and control needed to obtain clusters that are more aligned with the users' intent.  
We propose a new text-guided clustering approach named \modelname that uses an iterative discovery process,  guided by an unsupervised clustering objective, to generate interpretable visual concepts that better capture the criteria expressed in a user's instructions.  
We report superior performance compared to existing methods across a wide variety of image clustering and fine-grained classification benchmarks. 
\end{abstract}

\section{Introduction}
\vspace{-1.em}
Clustering is a fundamental task in computer vision, enabling the organization of unlabeled visual data into meaningful groups based on inherent patterns contained within. 
However, traditional clustering methods attempt to address an ill-posed problem, as for any given dataset, multiple equally different groupings/partitionings of the data can be valid. 
In the fully unsupervised setting, there is no guarantee that a clustering method effectively groups the data based on the visual criteria desired by the user.
To obtain groupings that encode structure that are relevant to the visual criteria of interest to users, some form of guidance to direct the clustering process is needed.  

Multiple  clustering (\ie multi-criteria) methods attempt to provide diverse partitionings of unlabeled data~\cite{hu2018subspace,wei2020multi,miklautz2020deep,metaxas2023divclust,yao2023augdmc}, but they cannot be guided based on users' preferences and any structure discovered needs to be explained post-hoc.  
Category discovery methods aim to group unlabeled data into clusters by leveraging the supervision signal available from  additional labeled data~\cite{han2019learning,vaze22generalized,wen2023simgcd}.
This labeled data guides the model during training to extract the types of visual features deemed important and relevant for defining clusters in the unlabeled data. 
This dependence on an already available labeled dataset limits the applicability of these methods in scenarios where such labels are unavailable or when the desired clustering criteria is dynamic and user-dependent. 
Moreover, existing methods can be biased towards fixed semantic labels (\eg semantic clusters~\cite{vaze2023no}), and lack the flexibility to accommodate multiple or nuanced clustering criteria. %

Recently a set of text-guided  clustering approaches have been proposed that capitalize on  advances in pre-trained large language models (LLMs) and vision-language models (VLMs). 
They enable users to specify the clustering criteria they wish to group images with via a concise high-level natural language instruction. 
IC$|$TC~\cite{kwon2024image} uses an VLM to caption images and then performs clustering in text space using an LLM, while Multi-MaP~\cite{yao2024multi} generates a set of language concepts using an LLM and then further refines VLM extracted image embeddings using an optimization approach. 
While effective, these methods can fail to best use the visual signal available~\cite{kwon2024image} or produce non-interpretable representations~\cite{yao2024multi}.  
Our work takes a different definition for interpretable clustering. Instead of generating a human readable cluster name like in IC$|$TC, our work performs clustering in a feature space where each feature is described by a semantic name. This approach enables a more fine-grained manipulation of the clustering by changing the features used for clustering.

We propose a novel \underline{I}terative and interpretable \underline{T}ext-\underline{G}uided image \underline{C}lustering approach named \textbf{\modelname} that is inspired by recent work on learning interpretable  image classifiers~\cite{chiquier2024evolving}. 
We begin with an Concept Generator, which processes a text query to produce an initial list of concepts relevant to a users' query. 
These language concepts are then mapped to images via a Concept Encoder which results in a set of embeddings that encodes the relevance of each concept for each image.
These concept vectors can be clustered to form groups. 
The quality of the resulting clusters is evaluated using an unsupervised loss which provides a feedback signal to the Concept Generator to guide the generation of a further refined set of concepts. 
This iterative process of concept generation, encoding, clustering, and evaluation continues until convergence is achieved, ensuring that the model learns to more effectively cluster images based on the user-provided criteria of interest. 
This allows for dynamic and flexible clustering without the need for labeled examples or predefined distance metrics. 
Our results show that \textbf{\modelname} obtains more accurate and meaningful clusterings,  outperforming existing zero-shot,  multiple clustering, category discovery, and text-guided clustering methods.

In summary, we make the following contributions:
(i) We introduce \textbf{\modelname}, a new approach for text-guided image clustering  that takes a text query as input and clusters unlabeled data using a learned interpretable feature space. 
(ii) We show that iterative application of our method, guided by an unsupervised objective, results in improved clusterings for successive time steps. 
(iii) We demonstrate superior performance over existing methods, providing interpretable and user-aligned results on both existing multi-criteria and fine-grained datasets.

\vspace{-1.5em}
\section{Related Work}
\vspace{-1.em}

\noindent{\bf Clustering.}   
Clustering is an unsupervised task where the goal is to partition data points into discrete groups. 
There is a large body of literature on classical methods that performing clustering on fixed input features~\cite{jain1999data}, but most success of late has come from deep clustering methods that also learn visual representations~\cite{ren2024deep}. 
Deep methods can be broadly grouped into those that perform some form of alternating optimization between cluster assignment and representation learning~\cite{xie2016unsupervised,yang2016joint} and those that perform end-to-end learning with a clustering specific objective~\cite{ji2019invariant,li2021contrastive,haeusser2019associative,huang2020deep}. 
Clustering objectives have also been shown to be effective for learning visual features in the context of representation learning~\cite{caron2018deep,asano2020self,caron2020unsupervised,van2020scan}. 
The main limitation of conventional clustering methods is that they only produce a single partitioning/clustering of the data. %

A single clustering cannot capture the full diversity of most datasets, where multiple valid groupings (e.g., by color or shape) may exist. Multi-criteria clustering addresses this by generating multiple partitions of the data~\cite{hu2018subspace}.
Note, multi-\emph{criteria} clustering is distinct from multi-\emph{view} clustering~\cite{yang2018multi}. 
In multi-view clustering each data instance is encoded using different modalities (\eg text and images).
Despite being more realistic, multi-criteria clustering remains underexplored, partly due to limited benchmark datasets. 
Basic methods rely on varying clustering hyperparameters~\cite{boongoen2018cluster}, but are constrained by feature expressiveness. 
Recent deep methods include matrix factorization requiring multi-view data~\cite{wei2020multi}, augmentation-based diversity~\cite{yao2023augdmc}, and loss functions promoting distinct clusterings~\cite{miklautz2020deep,metaxas2023divclust}. 
However, these remain fully unsupervised and lack user control over the clustering criteria.

\vspace{2pt}
\noindent{\bf Label-guided Clustering.} 
Given set of labeled and unlabeled data, category discovery methods aim to partition the unlabeled data into a set of discrete groups. 
They can can be viewed as a form of  label-guided clustering, whereby the labeled data informs they types of concepts/categories that should be discovered in the unlabeled data.  
For example, if the labeled data contained a set of birds species, it would be reasonable to assume that the unlabeled data should be grouped based on the `species' criteria. 
In novel category discovery it is assumed that the unlabeled data only contains instances from novel categories~\cite{han2019learning,Han2020automatically,han2021autonovel,zhao2021novel,Fini_2021_ICCV,Zhong_2021_CVPR,yang2023bootstrap}, while in the generalized setting the unlabeled data can contain instances from both previously seen and novel categories~\cite{vaze22generalized,cao22,wen2023simgcd,pu2023dynamic,fei2022xcon,zhang2022promptcal,vaze2023no,rastegar2024selex}. 
Various approaches have been explored in the literature, typically relying on some combination of self-supervised and supervised representation learning. %
There has also been work in other variants of the discovery problem such as online/incremental~\cite{zhao2023incremental,zhang2022grow,kim2023proxy,incd2022}, federated~\cite{pu2024federated}, active~\cite{ma2024active}, and work investigating the impact of different sets of labeled data~\cite{zhao2024labeled}. 

With limited exception, most previous  methods only investigate discovery in the context of one visual criteria, \ie the semantic category. 
This is predominately an artifact of how existing category discovery datasets are constructed, \ie the commonly used evaluation datasets are typically generated from existing visual categorization datasets~\cite{vaze22openset} where the categories contained within correspond to semantic concepts (\eg different dog breeds). 
As a result, there is only one type of visual criteria (\ie the semantic category) present in the data. 
In practice, image data can be grouped using many different visual criteria. 
This type of multi-criteria reasoning has been explored in metric learning in cases where the criteria are known in advance~\cite{veit2017conditional} or need to be estimated~\cite{kim2018context,nigam2019towards}. 
While these metric learning methods aim to learn representations that encode different visual criteria simultaneously, they do discover novel concepts within each criteria.  

\citet{vaze2023no} introduced the Clevr-4 dataset to study multi-criteria category discovery, extending CLEVR~\cite{johnson2017clevr} with four grouping criteria: shape, texture, color, and count. 
Their experiments show that pre-trained vision models are biased toward certain criteria (e.g., shape). 
However, their approach assumes access to labeled data for the target criterion. 
In contrast, we consider a more realistic setting where only the name of the desired visual criterion is known, without any labeled category examples.

\begin{figure*}
    \centering
    \includegraphics[width=0.9\linewidth]{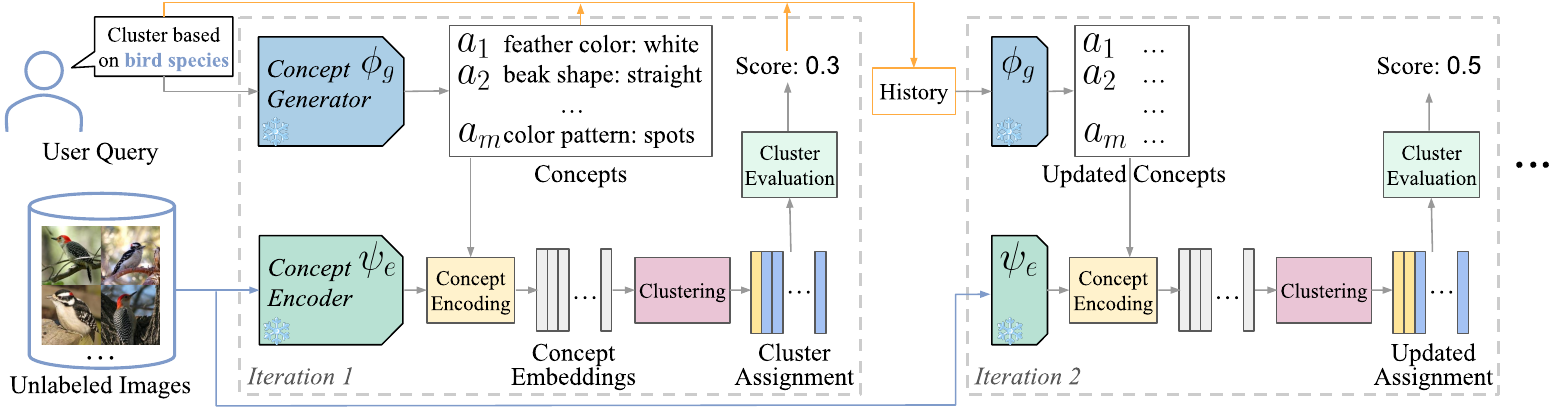} 
    \vspace{5pt}
    \caption{
    {\bf Overview of \modelname, our proposed text-guided image clustering method.}
    \textit{(Itr.~1)} The Concept Generator first generates an initial set of concepts based on the user provided text query. The Concept Encoder then encodes the images into the concept space to perform clustering.
    An unsupervised clustering evaluation score is used to measure how effective the clustering is.
    \textit{(Itr.~2)} Then the Concept Generator takes the entire previous history, including the concepts used for the previous clusterings and the scores for each clustering, to generate a new set of concepts.
    This iterative process is repeated until convergence, or a predefined number of iterations.
    }
    \label{fig:method}
    \vspace{-8pt}
\end{figure*}

\vspace{2pt}
\noindent{\bf Text-guided Clustering.} 
Recent advances in vision-language models (VLMs) enable zero-shot image classification by matching images to a predefined set of text-based concepts~\cite{radford2021learning,jia2021scaling,zhai2023sigmoid}. 
However, this requires the target concepts to be known in advance, limiting its use for clustering. To address this, \citet{li2024image} propose matching a large candidate vocabulary (e.g., WordNet nouns~\cite{miller1995wordnet}) to image clusters using CLIP, followed by filtering and k-means on combined text-image features. While effective for single-criteria clustering, their method is constrained by the vocabulary, produces only one clustering, and offers limited control.

The most relevant work to ours is IC$|$TC~\cite{kwon2024image}, which performs image clustering based on user-specified text criteria. 
Given a set of unlabeled images, a textual criterion, and the desired number of clusters, it uses a VLM (e.g., LLaVA~\cite{liu2024visual}) to caption each image, then summarizes the captions using an LLM (e.g., GPT-4~\cite{gpt4}) to generate candidate cluster labels. These are clustered via k-means, and the LLM assigns each caption to a cluster. Since the process operates solely on text after captioning, IC$|$TC is better viewed as a text clustering method grounded in image-generated captions.
Related work includes~\cite{stephan2024text}, which injects domain knowledge into the captioning step, and~\cite{liu2024organizing}, which jointly discovers both criteria and clusters from captions.
\cite{liu2024democratizing} also uses LLMs to generate concepts for fine-grained categorization, but does not address multi-criteria clustering where \modelname is focused on.
Additionally, in experiments we show that \modelname can improves the generated concepts via an unsupervised refinement process.

Multi-MaP~\cite{yao2024multi} similarly generates concept words with an LLM and optimizes their embeddings using gradient descent to match VLM features, but loses interpretability in the process of obtaining the continuous embeddings.
We also explore the same text-guided clustering setting as~\cite{kwon2024image,yao2024multi}, but introduce an iterative approach that refines concepts over time, resulting in an interpretable concept-bottleneck style~\cite{koh2020concept} representation for clustering.   

We take inspiration from~\cite{chiquier2024evolving}, which outlines an approach for evolving interpretable classifiers using a pre-trained LLM and VLM to iteratively discover a set of visual concepts which can be used to represent a class. 
Their goal is classification, and they are provided with images that are already grouped into different categories. 
However, the semantic category text labels are not provided.  They use the labels to train a classifier at each iteration to help find more relevant concepts. 
In contrast, we begin with an unlabeled set of images and are only provided a high-level criteria from the user which we use to discovery clusters.

\vspace{-15pt}
\section{Method}
\label{sec:instruct_clustering}
\vspace{-.5em}
We first introduce our text-guided image clustering problem setting, and then describe our new \textbf{\modelname} approach that leverages recent advances in vision-language and large language models (VLMs and LLMs) in an iterative framework.

\vspace{-1.em}
\subsection{Problem Formulation}
\label{sec:problem}
\vspace{-.5em}

Our goal is to cluster unlabeled images based on a user-defined criteria provided via a single natural language query/instruction. 
Our approach diverges from traditional clustering and category discovery methods by eliminating the need for labeled datasets or predefined distance metrics. 
Given a dataset $\mathcal{D} = \{\mathbf{x}_1, \mathbf{x}_2, \ldots, \mathbf{x}_n\}$, \ie a collection of images,  and a user provided natural language query $\mathcal{T}$, the objective is to partition $\mathcal{D}$ into $k$ clusters $\mathcal{C} = \{C_1, C_2, \ldots, C_k\}$ such that the clustering aligns with the visual criteria specified in the query.
$C_k=\{\mathbf{x}_{k_1}, \mathbf{x}_{k_2}, \ldots, \mathbf{x}_{k_n}\}$ contains the images in the $k$-th cluster.
Here, each image $\mathbf{x}_i$ is assigned to only one cluster, but more than one image can be assigned to each cluster as $k << n$. 
Thus we denote the clustering results as $\mathcal{D}_\mathcal{T}=\{(\mathbf{x}_1, y_1), (\mathbf{x}_2, y_2), \ldots, (\mathbf{x}_n, y_n)\}$, where $y_i \in \{1, 2, \ldots, k\}$ is the assignment for each image to a cluster.

\begin{figure*}[t]
\centering
\captionsetup{type=figure}
    \includegraphics[width=.80\linewidth]{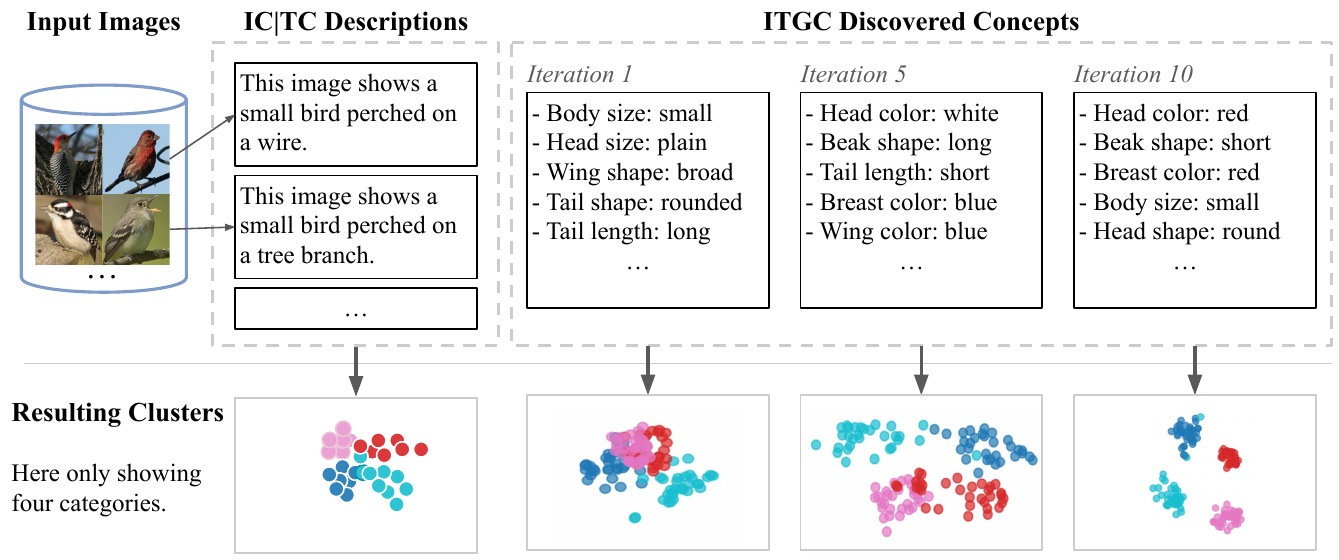}
    \vspace{5pt}
\caption{ 
\textbf{Examples of generated concepts for CUB.} IC$|$TC~\cite{kwon2024image} cannot meaningfully separate visual concepts when its captions for them are similar. 
In contrast,  \modelname discovers more interpretable concepts that better fit the clustering task, \ie bird species clustering.
}
\label{fig:qual_res}
\vspace{-6pt}
\end{figure*}

\vspace{-1.em}
\subsection{\modelname Overview}
\vspace{-.5em}
The core idea of \modelname  is to perform clustering in a language concept space, where each dimension represents a score indicating the presence of a single language-based concept in a given image. 
As a result, the clustering performed in this space can yield interpretable results.
By leveraging the knowledge encoded in LLMs, we can extract suitable concepts for the given query criteria to guide the clustering process.
In this context, a language concept refers to a specific value of a visual attribute present in an image. For example, instead of using a general concept like \emph{``bird wing colors''}, we specify it as \emph{``bird wing color: red''}.

Specifically, we utilize an LLM as a \textit{Concept Generator} $\phi_g$ to interpret the user provided query $\mathcal{T}$ and generate a set of relevant concepts/attributes $\mathcal{A} = \{a_1, a_2, \ldots, a_m\}$. 
The Concept Generator is explicitly prompted to generate concepts with specified values as mentioned previously.
These concepts serve as the basis for defining the clustering criteria.
Next, a \textit{Concept Encoder} $\psi_e$ is employed to project each data point $\mathbf{x}_i \in \mathcal{D}$ into the language concept space. 
This is achieved by encoding $\mathbf{x}_i$ into an embedding vector $\mathbf{z}_i$, where each dimension corresponds to the significance of a concept in $\mathcal{A}$. 
The resulting embeddings capture the semantic alignment of each data point with the user-defined criteria.
From this, an off-the-shelf clustering method such as k-means is applied to these embeddings to partition the dataset into clusters. 
This clustering process can be iteratively refined using `supervision' from an unsupervised clustering quality metric such as the silhouette score. 
As discussed later, this information provide feedback to the Concept Generator to enable it to  generate a more relevant set of concepts.
This iterative feedback loop continues until convergence of the unsupervised clustering quality metrics, better enabling  the clustering process to adhere to the user-defined query. 
By operating in the language concept space, \modelname provides enhanced flexibility and interpretability via a clustering that is closely aligned with human intent.

\cref{fig:method} illustrates an overview of \modelname.
The design of \modelname focuses on two key components: (i) effectively encoding images into the language concept space to capture the user-defined criteria and (ii) implementing an unsupervised iterative search process that refines clustering through continuous feedback and evaluation.
We present the details of these components in the following sections.

\vspace{-1.5em}
\subsection{Encoding Images into Concepts}
\label{sec:encode}
\vspace{-0.5em}

Encoding images into the concept space is a critical step in aligning the clustering process with the user-defined criteria. 
This involves transforming visual data into an interpretable representation that still reflects the user's criteria. 
We achieve this by leveraging pre-trained VLMs, \eg CLIP~\cite{radford2021learning} or LLaVA~\cite{liu2024visual}, to bridge the gap between vision and language.

For CLIP, the process begins by encoding each image $\mathbf{x}$ into the CLIP embedding space using its image encoder, $\mathbf{e}^{v}=\text{CLIP}_\text{image}(\mathbf{x})\in \mathbb{R}^d$, here $d$ is the latent dimension of CLIP. 
Simultaneously, the text descriptions of the LLM extracted concepts, generated by the Concept Generator, are encoded into the same embedding space using CLIP's text encoder $\mathbf{e}^{t_m}=\text{CLIP}_\text{text}(a_m)\in \mathbb{R}^d$. 
The similarity between the image embedding and each text embedding is then calculated, resulting in a vector $\mathbf{z}$, for each image, that denotes the relevance of each concept in an image. 
Specifically, the $m$-th dimension of $\mathbf{z}$ equals  the dot product of $\mathbf{e}^{v}$ and $\mathbf{e}^{t_m}$, \ie  $z_m = \mathbf{e}^{v} \cdot \mathbf{e}^{t_m}$.
This approach allows the clustering algorithm to operate in an interpretable space, aligned with the user's query,  facilitating more interpretable clustering.

In the case of LLaVA, the extraction of each concept is formulated as a question, such as \emph{``Please indicate the level of the presence of {concept} in the given image in a number between 0-10."} 
LLaVA processes these questions and provides numerical outputs that represent the presence of each concept in an image. 
These outputs are used to encode a vector for each image, capturing the semantic alignment with the user-defined criteria. 
This method enables a nuanced representation of concepts, enhancing the model's ability to adapt to diverse and complex datasets.

\vspace{-1.5em}
\subsection{Concept Refinement via Iterative Search}
\vspace{-0.5em}
The steps outlined previously result in a single set of language concepts and a set of per-image scores indicating how relevant the concepts are to each image. 
Next, inspired by~\cite{chiquier2024evolving}, we introduce an iterative search process designed to refine the clustering by continuously enhancing the alignment between the generated concepts and the user-defined criteria. 
A single instruction/prompt to the Concept Generator LLM may not yield sufficiently expressive queries, as the LLM lacks specific knowledge about which types of prompts best align with the Concept Encoder. 
To address this, we propose an iterative search mechanism. 
We note, that while our approach is iterative, it still only requires a single text query from the user at the start. 

At each iteration, the LLM, serving as the Concept Generator, has access to all previously generated sets of concepts and their corresponding evaluation scores. 
This historical context allows the LLM to assess the effectiveness of past concept sets and understand how the Concept Encoder responds to different prompts.
The LLM is explicitly instructed to mutate and evolve these previous concepts, where the aim is for it  to discover a set of concepts that yields improved unsupervised clustering performance. 
Importantly, the LLM is informed that its output will be used for encoding images, prompting it to generate concepts that correspond to visual features. 
This ensures that the generated concepts are interpretable and relevant for visual data.
The precise prompt used for mutating and evolving the previous concepts is provided in Appendix~\ref{appendix:prompts}.

The iterative process is facilitated by a feedback loop that evaluates the quality of the clusters using an supervised clustering metric that does not require label information. 
This feedback guides the LLM to adjust and refine the concept set $\mathcal{A}$, allowing it to better align with the Concept Encoder $\psi_e$ for the clustering task. 
Specifically, we denote the concept set used at the $j$-th iteration as $\mathcal{A}_j$, the feedback at the $j+1$-th iteration contains the concept sets $\{\mathcal{A}_1, \mathcal{A}_2, \ldots, \mathcal{A}_j\}$, and the silhouette scores for each of the concept sets $\{s_1, s_2, \ldots, s_j\}$, where $s_j$ is the silhouette score for the $j$-th iteration. 
The silhouette score is an unsupervised metric used to evaluate the quality of a clustering algorithm~\cite{silhouette}. 
By iteratively updating the concepts and re-projecting the images into the updated concept space, the \modelname converges towards a clustering solution that better satisfies the user's intent. 
This dynamic process not only enhances its adaptability to diverse datasets but also ensures that the clustering remains flexible and responsive to nuanced user preferences.
Note, that \modelname does not require any training of the vision backbone model or the language model, as the search iterations are purely done in the language space using prompts. 
Thus, the search process of our model can be completed much faster compared to other methods that use gradient-based optimization~\cite{yao2024multi,yao2023augdmc}.
The overall pseudo-code of our method is provided in the appendix.

\vspace{-1.5em}
\section{Experiments}

\vspace{-1.em}
\subsection{Implementation Details}
\vspace{-0.5em}
\label{sec:implement}

\noindent{\bf Models.}
We utilize existing pre-trained large models where possible.
Specifically, for the Concept Generator, we use the open-sourced LLM from DeepSeek-V2~\cite{liu2024deepseek}, as it demonstrates strong performance at a relatively low cost. 
We also include results using the closed-source GPT-4~\cite{gpt4} which further enhances the performance.
We evaluate two different ways of encoding images into the concept space for our Concept Encoder; a CLIP or a LLaVA encoder. 
For the CLIP model, we use the CLIP-L/14 model, and for the LLaVA model, we use the LLaVA-v1.6-34B model. 
For the clustering process, we perform k-means clustering. 
Note, here we assume the number of clusters $k$ is known before hand for us to perform k-means clustering.
However, it is straightforward to use other clustering methods such as DBSCAN~\cite{ester1996density} which do not require the number of clusters  a-priori (see Appendix). %

\vspace{2pt}
\noindent{\bf Datasets.}
We evaluate our method on a diverse set of datasets to showcase its versatility. 
Specifically, two different dataset types are used; multi-criteria and fine-grained classification datasets. 
The multi-criteria clustering datasets each contain between two and four different high-level visual criteria, where each criteria induces a different clustering of the data.  
For example, for the criteria  `shape', objects are grouped based on their shape, whereas the criteria `color' results in a different grouping. 
We evaluate on the Stanford 40 Actions~\cite{yao2011human}, ImageNet-R~\cite{hendrycks2021many}, Fruits-360~\cite{fruits360}, Cards~\cite{cards}, and the larger-scaled dataset Clevr-4~\cite{vaze2023no}. 
These datasets present various challenges in discovering multiple independent clusterings structures within the datasets. 
For fine-grained evaluation, we use CUB-200~\cite{cub200}, Stanford Cars~\cite{stanfordcars}, and FGVC Aircraft~\cite{aircraft}. %
The fine-grained nature of these datasets necessitates that a model can understand nuanced visual concepts and thus represent a challenging test case of our ability to generate interpretable spaces for clustering.

\vspace{2pt}
\noindent{\bf Evaluation.}
The evaluation protocol is similar to that used in category discovery~\cite{vaze22generalized}, as we need to generate a match between the predicted clusterings from the model and the ground truth clustering.
During evaluation, given the ground truth $y^*$ and the predicted label $\hat{y}$, the accuracy (ACC) is calculated as $\text{ACC} = \frac{1}{M} \sum_{i=1}^{M} \mathds{1}(y^*_i = p(\hat{y}_i))$ where $M = |\mathcal{D}|$, and $p$ is the optimal permutation that matches the predicted cluster assignments to the ground truth.

\begin{table*}[t]
    \centering
    \resizebox{0.92\linewidth}{!}{
    \begin{tabular}{l |l| ccc|cccc|cc}
    \toprule
    Dataset & Type & \multicolumn{3}{c|}{Stanford 40 Actions~\cite{yao2011human}} & \multicolumn{4}{c|}{Clevr-4~\cite{vaze2023no}} & \multicolumn{2}{c}{ImageNet-R~\cite{hendrycks2021many}} \\%& \multicolumn{2}{c}{Fruits-360} & \multicolumn{2}{c}{Cards} \\
    \midrule
    Criteria & & Action &  Location & Mood & Texture & Shape & Count & Color& Class  & Rendition \\%& Species  & Color   &  Rank   & Suits \\
    \midrule
    CLIP~\cite{radford2021learning}$^*$ & Zero-shot    & 78.3 & 80.4 & 80.1 & 34.2 & 74.5 & 74.5 & 33.5 & 75.8 & 33.5 \\
    SimGCD~\cite{wen2023simgcd}$^\dagger$     & Discovery & 78.6 & 79.4 & 80.4 & 35.6 & 75.0 & 74.6 & 34.0 & 76.3 & 33.8 \\ \hline
    SCAN~\cite{van2020scan}         & Clustering   & 39.7 & 35.9 & 25.0 & 28.5 & 54.1 & 30.6 & 31.5 & 60.1 & 18.9 \\
    AugDMC~\cite{yao2023augdmc}     & Clustering & 65.6 & 45.6 & 40.2 & 28.9 & 50.2 & 35.6 & 27.6 & 45.3 & 25.4 \\
    IC$|$TC~\cite{kwon2024image}    & Text-guided & 77.4 & 82.2 & 79.3 & 37.9 & 73.5 & 76.5 & 30.2 & 69.3 & 31.9 \\
    Multi-MaP~\cite{yao2024multi}   & Text-guided& 79.5 & 83.5 & 81.5 & 34.5 & 75.6 & 74.8 & 35.6 & 76.4 & 34.0 \\
    \midrule
    Ours (w.~LLaVA~\cite{liu2024visual})  & Text-guided &  81.0 & 81.2 & 83.5  & 36.0 & 76.0 & 75.2 & 36.2 & 77.0 & 33.6 \\
    Ours (w.~CLIP~\cite{radford2021learning}) & Text-guided &  80.2 & 86.7 & 85.3  & 35.3 & 76.4 & 75.7 & 36.0 & 78.4 & 36.7 \\
    \bottomrule
    \end{tabular}
    }
    \vspace{5pt}
    \caption{
    \textbf{Results on multi-clustering datasets.}
    Our \modelname approach demonstrates superior performance compared to existing baselines.
    The zero-shot CLIP performance denoted with~$^*$ has access to the ground truth class names, and the SimGCD method denoted with $^\dagger$ has access to a small set of labeled examples.
    We can see that our method obtains stronger performance   than these two methods despite their use of this additional information, indicating the effectiveness of our iterative search process.
    }
    \label{tab:comp_mc} 
    \vspace{-8pt}
\end{table*}

\vspace{-12pt}

\subsection{Results on Multi-clustering Datasets}
\vspace{-0.5em}
In \cref{tab:comp_mc} we evaluate \modelname on several  multi-clustering datasets. %
Compared to the baseline methods of zero-shot CLIP, which receives the cluster names as input, and SimGCD, which is provided with a set of labeled examples as input, we observe that \modelname  consistently outperforms them  under the different input criteria/queries.
This indicates the effectiveness of our method in understanding the clustering criteria.
Compared to methods that are specifically designed to handle text-guided and multi-criteria clustering (\eg AugDMC, IC$|$TC, and Multi-MaP), we obtain an increase in performance. 
We attribute this performance improvement  to the better alignment of the language concepts and the concept encoder. 
This concept alignment is a result of the iterative search procedure of our method, which enables it to automatically find the concepts more relevant to the user indicated clustering task. 
Previous methods either use one-off prompting~\cite{radford2021learning}, hand-designed augmentation priors~\cite{yao2023augdmc,yao2024multi}, or human-in-the-loop optimizations~\cite{kwon2024image} that do not fully utilize the possible space of concepts for the clustering.
Fruits360 and Cards are two datasets that have been used in previous literature on multi-criteria clustering. 
In \cref{tab:easier_mc} we report performance on these two datasets. 
We consider Fruits360 and Cards as easy datasets as the criteria for clustering is simple and the number of images in the datasets is also small. 
We can see that all models perform well on these two datasets, and that our proposed approach also obtains strong performance across the different clustering criteria.

\begin{table*}[t]
\centering
\begin{minipage}{0.49\textwidth}
\centering
\resizebox{\columnwidth}{!}{
\begin{tabular}{l |l| cc | cc}
\toprule
Dataset & Type & \multicolumn{2}{c|}{Fruits-360~\cite{fruits360}} & \multicolumn{2}{c}{Cards~\cite{cards}} \\
\midrule
Criteria & & Species & Color & Rank & Suits \\
\midrule
CLIP~\cite{radford2021learning}$^*$ & Zero-shot & 92.5 & 91.4 & 96.7 & 93.0 \\
SimGCD~\cite{wen2023simgcd}$^\dagger$ & Discovery & 90.1 & 91.9 & 95.0 & 93.4 \\ \hline
AugDMC~\cite{yao2023augdmc} & Clustering & 89.7 & 90.5 & 87.3 & 80.6 \\
IC$|$TC~\cite{kwon2024image} & Text-guided & 90.4 & 89.2 & 93.4 & 92.5 \\
Multi-MaP~\cite{yao2024multi} & Text-guided & 90.8 & 89.8 & 93.2 & 91.4 \\
\midrule
Ours (w.~LLaVA~\cite{liu2024visual}) & Text-guided & 90.4 & 90.5 & 94.6 & 91.0 \\
Ours (w.~CLIP~\cite{radford2021learning}) & Text-guided & 91.6 & 90.1 & 95.8 & 93.7 \\
\bottomrule
\end{tabular}
}
\vspace{5pt}
\caption{\textbf{Results on additional multi-criteria clustering datasets.} Here, $^*$~denotes access to ground truth class name and $^\dagger$~denotes access to a small set of labeled examples.}
\label{tab:easier_mc}
\end{minipage}
\hfill
\begin{minipage}{0.49\textwidth}
\centering
\vspace{1em}
\resizebox{\columnwidth}{!}{
\begin{tabular}{l |l| ccc}
\toprule
Dataset & Type & CUB & Cars & Aircraft \\
\midrule
CLIP~\cite{radford2021learning}$^*$ & Zero-shot & 51.4 & 60.1 & 25.6 \\
SimGCD~\cite{wen2023simgcd}$^\dagger$ & Discovery & 53.4 & 61.3 & 27.9 \\ \hline
AugDMC~\cite{yao2023augdmc} & Clustering & 24.6 & 35.6 & 20.5 \\
IC$|$TC~\cite{kwon2024image} & Text-guided & 48.9 & 62.1 & 31.4 \\
Multi-MaP~\cite{yao2024multi} & Text-guided & 47.8 & 64.0 & 32.1 \\
\midrule
Ours (w.~LLaVA~\cite{liu2024visual}) & Text-guided & 52.7 & 63.5 & 28.7 \\
Ours (w.~CLIP~\cite{radford2021learning}) & Text-guided & 57.8 & 65.4 & 34.5 \\
\bottomrule
\end{tabular}
}
\vspace{5pt}
\caption{\textbf{Results on fine-grained classification datasets.} These fine-grained datasets require models to optimize for subtle visual details to separate the categories. Our \modelname approach shows significant improvements, discovering sufficient fine-grained structure.}
\label{tab:fg}
\end{minipage}
\vspace{-.5em}
\end{table*}

\vspace{-1.0em}
\subsection{Results on Fine-grained Datasets}
\vspace{-0.5em}
Here we evaluate our method on fine-grained classification datasets which  only contain a single clustering criteria (\ie the semantic category). 
This requires models to be able to capture and compare the subtle visual difference within the image clusters to be able to distinguish fine-grained classes. 
We report performance on CUB-200~\cite{cub200}, Stanford Cars~\cite{stanfordcars}, and FGVC-Aircraft~\cite{aircraft} in~\cref{tab:fg}.
We can see that these  datasets pose a greater challenge for the models evaluated. 
This is evident by the fact that the CLIP baseline does not perform well on these datasets, despite being given the class names. 
SimGCD, a category discovery method, leverages a set of labeled examples, and thus demonstrates  improved performance.
If the signal for learning the clustering criteria is only applied on the coarse augmentation level, as in the case of  AugDMC,  we observe that it does learn to perform fine-grained clustering well. 
IC$|$TC and Multi-MaP also demonstrate a lower performance as they only optimize the search for the suitable clustering criteria in the text space.
Our method demonstrate strong performance on these  datasets, as it is able to better align the language concept space to the encoder space. 
In \cref{fig:qual_res} we display qualitative results comparing the clustering results of our \modelname approach to the recent IC$|$TC~\cite{kwon2024image} on CUB.
In contrast, \modelname is able to generate more detailed concepts that are relevant to the objects of interest in the images.

\vspace{-1.5em}
\subsection{Ablations}
\vspace{-.5em}

\begin{wraptable}{r}{0.5\textwidth}
    \centering
    \resizebox{0.5\textwidth}{!}{
    \begin{tabular}{l c |cccc| c}
    \toprule
    Encoder & \# Itrs. & \multicolumn{4}{c|}{Clevr-4} & CUB \\
    \midrule
    Criteria &   &  Texture & Shape & Count & Color & Species \\
    \midrule
    CLIP   & 1  &   31.5 & 70.1 & 71.5 & 30.0 & 46.7 \\
    CLIP   & 5  &   34.2 & 72.6 & 73.8 & 33.5 & 50.3 \\
    CLIP   & 10 &   35.7 & 75.9 & 75.4 & 35.8 & 57.9 \\
    CLIP   & 15 &   35.3 & 76.4 & 75.7 & 36.0 & 57.8 \\
    \midrule
    LLaVA   & 1  &  30.1 & 68.2 & 66.2 & 27.8 & 43.5 \\
    LLaVA   & 5  &  32.4 & 70.1 & 68.3 & 31.2 & 44.3 \\
    LLaVA   & 10 &  35.1 & 74.5 & 73.5 & 34.5 & 48.6 \\
    LLaVA   & 15 &  36.0 & 76.0 & 75.2 & 36.2 & 52.7  \\
    \bottomrule
    \end{tabular}
    }
    \vspace{5pt}
    \caption{\textbf{Impact of search iterations.}}
    \label{tab:iteration}
    \vspace{-.4cm}
\end{wraptable}

\vspace{2pt}
\noindent{\bf Search Iterations.}
In~\cref{tab:iteration}, we present the results of varying the number of search iterations.
At the beginning of the search, the \modelname already results in a good clustering thanks to the image-text alignment ability of the CLIP (or LLaVA) model and the reasoning ability of the LLM.
As we increase the number of search iterations, the Concept Generator begins to align more effectively with  the Concept Encoder, and therefore results in improved performance.
We can also see that for CLIP the performance saturates after a certain number of iterations, this indicates the search converges to a set of useful concepts  for the given clustering criteria.
The search process only involves prompting an LLM and performing forward passes through a pre-trained model. 
As a result, the computational cost of our search process is significantly lower than gradient-based optimization methods.

\vspace{2pt}
\noindent{\bf Stronger Models.}
In~\cref{tab:strong_models}, we explore the impact of using stronger models for both the Concept  Generator and Encoder in \modelname.
CoCa~\cite{Yu2022CoCaCC} improves upon CLIP~\cite{radford2021learning} by adding an additional image captioning objective to enable the model to obtain better alignment, and EVA-CLIP~\cite{EVA-CLIP} incorporates improved representation learning, optimization, and augmentation to achieve superior performance compared to previous CLIP models with a smaller training cost.
Using stronger models than CLIP for the Concept Encoder improves performance. 
This demonstrates that the iterative search process in \modelname is able to utilize the better alignment of the image and text from the stronger models.
LLaVA-OneVision~\cite{li2024llava} and GPT-4o~\cite{openai2024gpt4o} are more performant than the LLaVA-1.6 model used in previous experiments.
LLaVA-OneVision utilizes more instruction tuning data to fine-tune the model and supports higher resolution inputs. 
It has also uses a larger LLM, QWen-2.0 72B~\cite{qwen},  as its decoder.
GPT-4o~\cite{openai2024gpt4o} is one of the most advanced multi-modal language model currently available for image understanding tasks.
Our evaluation  utilizing these stronger models  demonstrates that \modelname is able to leverage the stronger models' capabilities to perform superior clustering that better matches the user's intent.

\begin{table*}[t]
\centering
\begin{minipage}{0.48\textwidth}
\centering
\resizebox{\columnwidth}{!}{
\begin{tabular}{l | cccc | c}
\toprule
Encoder &  \multicolumn{4}{c|}{Clevr-4} & CUB \\
\midrule
Criteria & Texture & Shape & Count & Color & Species \\
\midrule
CLIP~\cite{radford2021learning} & 35.3 & 76.4 & 75.7 & 36.0 & 57.8 \\
CoCa~\cite{Yu2022CoCaCC} & 36.7 & 77.8 & 76.2 & 36.5 & 59.3 \\
EVA-CLIP~\cite{EVA-CLIP} & 37.2 & 78.4 & 76.8 & 37.5 & 61.0 \\
\midrule
LLaVA-1.6~\cite{liu2024visual} & 36.0 & 76.0 & 75.2 & 36.2 & 52.7 \\
LLaVA-OneVision~\cite{li2024llava} & 37.6 & 77.2 & 75.4 & 36.8 & 53.2 \\
GPT-4o~\cite{openai2024gpt4o} & 38.9 & 79.4 & 77.3 & 37.5 & 57.6 \\
\bottomrule
\end{tabular}
}
\vspace{5pt}
\caption{\textbf{Impact of stronger models.}}
\label{tab:strong_models}
\end{minipage}
\hfill
\begin{minipage}{0.5\textwidth}
\centering
\resizebox{\columnwidth}{!}{
\begin{tabular}{l | l | cccc | c}
\toprule
Encoder & Auxiliary Info. & \multicolumn{4}{c|}{Clevr-4} & CUB \\
\midrule
Criteria & & Texture & Shape & Count & Color & Species \\
\midrule
CLIP & None & 35.3 & 76.4 & 75.7 & 36.0 & 57.8 \\
CLIP & Partial Labels & 38.4 & 80.0 & 79.4 & 38.5 & 59.9 \\
CLIP & Example Class Name & 38.6 & 80.4 & 79.2 & 38.7 & 59.7 \\
\midrule
LLaVA & None & 36.0 & 76.0 & 75.2 & 36.2 & 52.7 \\
LLaVA & Partial Labels & 37.1 & 77.9 & 78.4 & 37.5 & 53.4 \\
LLaVA & Example Class Name & 38.0 & 78.4 & 79.0 & 37.6 & 53.3 \\
\bottomrule
\end{tabular}
}
\vspace{5pt}
\caption{\textbf{Impact of auxiliary information.}}
\label{tab:aux_info}
\end{minipage}
\vspace{-1.em}
\end{table*}

\vspace{2pt}
\noindent{\bf Using Auxiliary Information for Search.}
In our previous experiments, the search process is performed using only the feedback from the silhouette score derived from the current clustering. 
In~\cref{tab:aux_info}, we present results where we use richer feedback signals to further assist the search process.
`Partial Labels' denotes the case where we allow the model to have access to the labels of 10\% of all examples for the specified clustering criteria. %
The model can use the clustering accuracy on these examples as a more direct feedback signal for the iterative search process.
`Example Class Name' adds three example ground truth class names, for the given dataset and criteria,  to the prompt for the Concept Generator so it can produce more relevant concepts for clustering.
Note that these two forms of auxiliary information can be relatively cheap to obtain in practice, yet from our experiments, can greatly improve the performance of \modelname. 
This illustrates the flexibility of \modelname, demonstrating that it performs well given varying amounts of supervision.

\vspace{-1.7em}
\section{Conclusion}
\vspace{-1.em}
We introduced \modelname, a new approach for text-guided clustering  that uses an  iterative search procedure to discover interpretable concepts for performing clustering to satisfy a user provided criteria.
The  search procedure is guided by an unsupervised clustering objective, and can generate interpretable concepts to help the user understand the clustering results. 
In our experiments, we demonstrate that \modelname is able to outperform existing baselines on the tasks of multi-criteria clustering as well as fine-grained image clustering.

\noindent \textbf{Limitations}.
Despite our performance improvements over previous methods on the challenging real-world problem of text-guided clustering, limitations still exist.
For example not all visual concepts can be conveyed easily via language, thus \modelname will not work as well in such cases.
For some fine-grained concepts, visual examples may need to be used, which is not possible with \modelname.

\noindent {\bf Acknowledgements.}  This work was in part supported by a Royal Society Research Grant.

\bibliography{bmvc_final}

\clearpage
\pagebreak

\appendix
\setcounter{table}{0}
\renewcommand{\thetable}{A\arabic{table}}
\setcounter{figure}{0}
\renewcommand{\thefigure}{A\arabic{figure}}

\noindent{\huge Appendix}

\section{Additional Results}

\subsection{LLaVA Logits}
\label{sec:llava_logits}
Here we explore an alternative way of utilizing the LLaVA~\cite{li2024llava} model for the Concept Encoder $\psi_e$.
The approach described in the main paper directly prompts  LLaVA to output a number/score to indicate the level of presence of a certain concept in the image.
Recent work has shown that simply prompting a language model for such scores can be unreliable~\cite{tian2023just}.
Thus, here we take an alternative approach, and prompt LLaVA  with a binary choice question: ``Does the image contains the concept of {concept}? A. Yes, B. No'', and then we take the logit score of the token `A' as the encoded value for the concept.

\cref{tab:llava_logits} compares this alternative approach to the one we used in the main paper and to using CLIP as the Concept Encoder $\psi_e$. 
We observe that using the logit score of LLaVA yields better results than using the text output.  
However, the performance of LLaVA with the logit score is still lower than our method using CLIP. Considering the number of parameters in LLaVA compared to CLIP, this indicates that more exploration need to be done to investigate how to best leverage vision language models in order to understand diverse visual concepts.

\subsection{Impact of Clustering Algorithm}
In~Sec.~4.1 in the main paper, 
we noted that it is possible to use clustering algorithms other than k-means to perform the clustering in the space encoded by the Concept Encoder $\psi_e$.
\cref{tab:c_alg} demonstrates the results of using different clustering algorithms.
We specifically compare with the DBSCAN algorithm which does not require the number of clusters to be known a-priori.
We can see that compared to the k-means baseline, DBSCAN demonstrates a lower performance. 
However, by leveraging a hierarchical variant of it (\ie H-DBSCAN), we observe a improvement, where the performance is comparable to the k-means baseline which assumes that the number of clusters is known.
These results indicates that our framework works \emph{without} knowledge of the number of clusters a-priori.

\begin{table*}[h]
\centering
\begin{minipage}{0.49\textwidth}
\centering
\resizebox{\columnwidth}{!}{
\begin{tabular}{l l |cccc| c}
\toprule
Encoder & Score & \multicolumn{4}{c|}{Clevr-4} & CUB \\
\midrule
& & Texture & Shape & Count & Color & Species \\
\midrule
LLaVA & Prompts & 32.4 & 70.1 & 68.3 & 31.2 & 44.3 \\
LLaVA & Logits & 34.3 & 72.7 & 69.8 & 33.7 & 48.9 \\
CLIP & Matching & 35.3 & 76.4 & 75.7 & 36.0 & 57.8 \\
\bottomrule
\end{tabular}
}
\vspace{5pt}
\caption{\textbf{Impact of utilizing different scoring functions for encoding concepts.}}
\label{tab:llava_logits}
\end{minipage}
\hfill
\begin{minipage}{0.49\textwidth}
\centering
\resizebox{\columnwidth}{!}{
\begin{tabular}{l l |cccc| c}
\toprule
Encoder & Algorithm & \multicolumn{4}{c|}{Clevr-4} & CUB \\
\midrule
& & Texture & Shape & Count & Color & Species \\
\midrule
CLIP & k-means & 35.3 & 76.4 & 75.7 & 36.0 & 57.8 \\
CLIP & DBSCAN & 34.7 & 73.5 & 74.2 & 33.7 & 54.7 \\
CLIP & H-DBSCAN & 35.0 & 75.9 & 75.5 & 35.6 & 56.9 \\
\bottomrule
\end{tabular}
}
\vspace{5pt}
\caption{\textbf{Impact of utilizing different clustering algorithms.}}
\label{tab:c_alg}
\end{minipage}
\end{table*}

\section{Qualitative Results}

\begin{figure*}[h]
    \centering
    \includegraphics[width=0.95\linewidth]{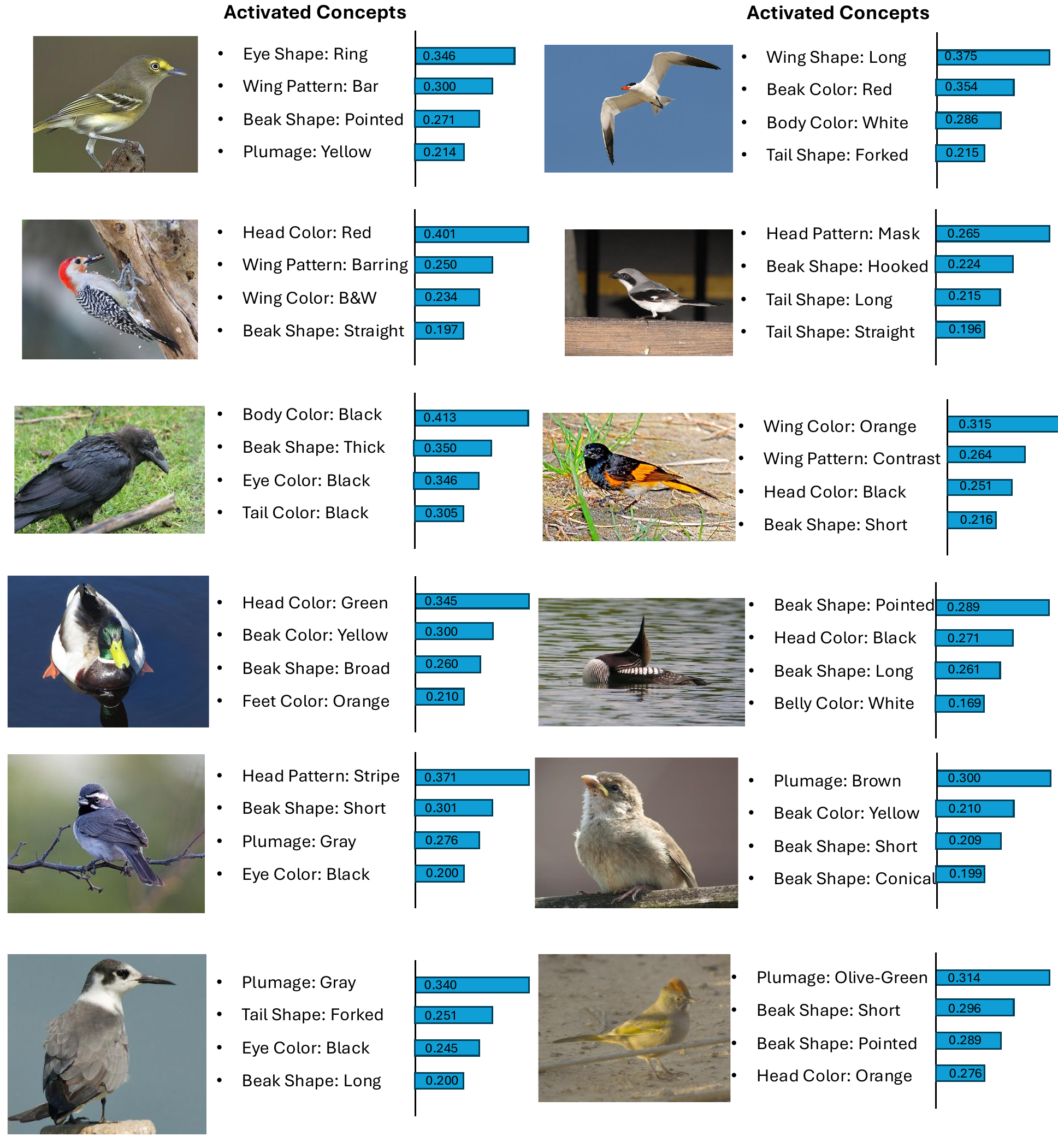}
    \vspace{5pt}
    \caption{\textbf{Discovered concepts on images from the CUB dataset.} The bars shows the cosine similarity of the image and the corresponding text concept in the CLIP feature space.}
    \label{fig:concept_image_example}
\end{figure*}

Here, we present qualitative results of the outputs obtained from \modelname.  
In \cref{fig:concept_image_example}, we randomly sample several images from the CUB dataset and plot the top-4 highly activated text concepts.
From these examples, it is clear that \modelname is able to discover the most salient concepts within a dataset.
Besides the visual examples presented in this file, we have also included a list of the final discovered attributes as part of the supplementary material.\footnote{The list of attribute can be accessed from the `cub\_concepts.txt' file in the same zip file.}

\section{Additional Implementation Details}

\subsection{Datasets}
\cref{tab:datasets} provides a summary of the datasets used in our experimental evaluation. 
It contains a mix of coarse and fine-grained visual datasets.

\begin{table}[h]
    \centering
    \resizebox{.5\textwidth}{!}{
    \begin{tabular}{l ccc}
    \toprule
    Dataset     &  \# criteria & \# classes & \# images\\
    \midrule
    Stanford 40 Actions~\cite{yao2011human}   &   3    &   40; 10; 4         &  9.5k  \\
    Clevr-4~\cite{vaze2023no}                 &   4    &   10; 10; 10; 10         &  100k \\
    ImageNet-R~\cite{hendrycks2021many}       &   2    &   200; 16         &  30k \\
    Fruits-360~\cite{fruits360}               &   2    &   4; 4         &  4.8k \\
    Cards~\cite{cards}                        &   2    &   13; 4         &  8.0k \\
    \midrule
    CUB-200~\cite{cub200}                     &   1    &    200     & 6.0k \\
    Stanford Cars~\cite{stanfordcars}         &   1    &    198     & 6.1k  \\
    FGVC-Aircraft~\cite{aircraft}             &   1    &    102     & 6.7k \\
    \bottomrule
    \end{tabular}
    }
    \vspace{5pt}
    \caption{\textbf{Evaluation datasets.} Multi-criteria datasets are those containing more than one annotated criteria.}
    \label{tab:datasets}
\end{table}

\subsection{Pseudo-code}

Alg.~\ref{algo:code} outlines the main steps in  \modelname in the form of pseudo-code. 

\begin{algorithm}[t]
\small
\SetAlgoLined
\PyCode{def clustering(dataset, user\_query):}\\
\Indp
  \PyComment{Clustering following user query}\\
  \PyCode{history = []} \\
  \PyCode{while not converged:} \\
  \Indp
    \PyComment{Generate concept for clustering based user prompt and history.}\\
    \PyCode{concepts = concept\_generator(} \\
    \Indp
      \PyCode{user\_prompt,} \\
      \PyCode{history} \\
    \Indm
    \PyCode{)}\\
    \PyCode{proj\_dataset = []} \\
    \PyCode{for data in dataset:} \\
    \Indp
      \PyCode{proj\_concept = concept\_encoder(}\\
      \Indp
          \PyCode{data,} \\
          \PyCode{concepts}\\
      \Indm
      \PyCode{)}\\
      \PyCode{proj\_dataset += [proj\_concept]}\\
    \Indm
    \PyComment{Run clustering on the concept space}\\
    \PyCode{cluster = kmeans(proj\_dataset)}\\
    \PyCode{score = silhouette(cluster)} \\
    \PyCode{history += [(concepts, score)]}\\
  \Indm
  \PyCode{return concepts}
\caption{Pseudo-code for \modelname}
\label{algo:code}
\end{algorithm}

\subsection{Prompts}
\label{appendix:prompts}
Here we outline the text prompts we used in our experiments.
These include the prompt used to mutate and evolve based on previous concepts generated and the prompt used to inform the model about the task and generate the first set of concepts.

First we present the prompt used to inform the model about the task and generate the first set of concepts: 
\begin{tcolorbox}[colback=promptcolor, boxrule=0.5pt, arc=2mm, boxsep=0mm, title=Task and initial concept generation, colbacktitle=lightgray, coltitle=black, toptitle=0.5mm, bottomtitle=0.5mm, fonttitle=\bfseries]
You are an expert in generating visual concepts that can be used to classify visual categories. 
Now you have a dataset and you want to generate visual concepts that are useful for classifying the \{CRITERIA\} of the given dataset.
Please start each line of attributes with `-` and only one attribute per line.
Please generate statements containing specified values for the attribute, for example, generate ``color: red", instead of ``color".
\end{tcolorbox}
In this prompt, ``\{CRITERIA\}" is a placeholder that will be replaced by the value of the actual user defined criteria/query.

Next is the prompt for mutating and evolving the previous concepts.
This prompt is used for prompting the Concept Generator $\phi_g$ to generate a set of new concepts for the Concept Encoder $\psi_e$ to encode into the embedding space for further clustering: 
\begin{tcolorbox}[colback=promptcolor, boxrule=0.5pt, arc=2mm, boxsep=0mm, title=Additional concept generation, colbacktitle=lightgray, coltitle=black, toptitle=0.5mm, bottomtitle=0.5mm, fonttitle=\bfseries]
Here are sets of attributes that is used as sets of descriptors for classifying a set of images into different classes. 
These attributes are used for the classification on the class criteria of \{CRITERIA\}.
Each sets of attributes are ranked according to the silhouette score for clustering the data.
Based on what you've seen below, propose a new set of visual attributes, that you think might achieve an even higher score. 
Please start each line of attributes with `-` and only one attribute per line.
Please generate statements containing specified values for the attribute, for example, generate ``color: red", instead of ``color".
The following lines contains the previously used sets of attributes and their score, mutate and evolve upon them to generate a new set of attributes to achieve a higher score:\\
\{Historical attributes 1\} \{score 1\}\\
\{Historical attributes 2\} \{score 2\}\\
...\\
\{Historical attributes n\} \{score n\}\\
Now please generate a new set of attributes based on the previous instructions.
\end{tcolorbox}

Different from the prompts used in~\cite{chiquier2024evolving}, our prompt emphasizes generating a set of concepts that can be used to form a space for clustering different classes, while~\cite{chiquier2024evolving} emphasizes on generating a set of attributes that can be used to describe one novel class.

We use the following text to prompt LLaVA to obtain a text-based score for concept encoding: 
\begin{tcolorbox}[colback=promptcolor, boxrule=0.5pt, arc=2mm, boxsep=0mm, title=LLaVA prompt 1, colbacktitle=lightgray, coltitle=black, toptitle=0.5mm, bottomtitle=0.5mm, fonttitle=\bfseries]
Please indicate the level of the presence of \{concept\} in the given image in a number between 0-10.
\end{tcolorbox}

Additionally, as show in~\cref{sec:llava_logits},  we can use the logits from LLaVA for the concept encoding score. 
In this case, we use the following prompt:
\begin{tcolorbox}[colback=promptcolor, boxrule=0.5pt, arc=2mm, boxsep=0mm, title=LLaVA prompt 2, colbacktitle=lightgray, coltitle=black, toptitle=0.5mm, bottomtitle=0.5mm, fonttitle=\bfseries]
 Does the image contains the concept of \{concept\}? A. Yes, B. No
\end{tcolorbox}
We take the logits of the token ``A" as the encoded concept score.

\begin{figure*}[h]
    \centering
    \includegraphics[width=0.95\linewidth]{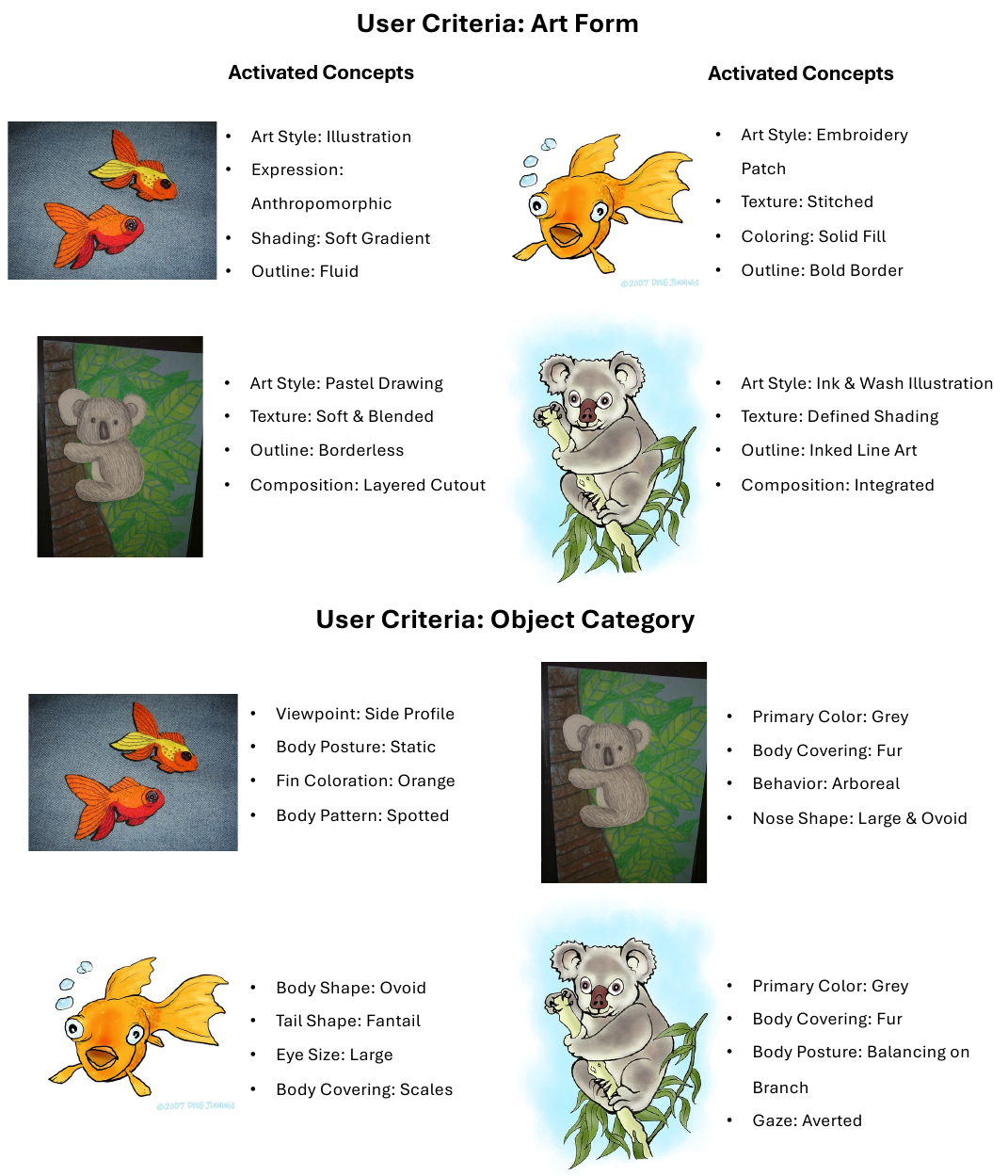}
    \caption{\textbf{Discovered concepts on images from ImageNet-R dataset with different user provided criteria.} Only showing the top-4 activated concepts.}
\end{figure*}

\end{document}